\documentclass[10pt,conference]{IEEEtran}

\usepackage[font=footnotesize]{caption}
\usepackage{subcaption}
\usepackage{amsmath,amssymb,amsfonts}
\usepackage{siunitx}
\usepackage{nicefrac}
\usepackage{graphicx}
\usepackage{textcomp}
\usepackage{xcolor}
\usepackage{algorithmic}
\usepackage{algorithm}
\usepackage{array}
\usepackage{hyperref}
\usepackage{multirow}
\usepackage{url}
\usepackage{verbatim}
\usepackage{cite}
\usepackage{booktabs}
\usepackage{balance}
\usepackage[nolist,printonlyused]{acronym}
\usepackage[capitalize]{cleveref}
\usepackage[normalem]{ulem}

\begin{document}

\title{Practical Policy Distillation for Reinforcement Learning in Radio Access Networks}

\author{
\IEEEauthorblockN{Sara Khosravi, Burak Demirel, Linghui Zhou, Javier Rasines, Pablo Soldati}
\IEEEauthorblockA{\textit{Ericsson AB, Kista, Sweden} \\
email:\{name.surname\}@ericsson.com}
}

\maketitle

\begin{figure}[!b]
\centering
\fbox{\parbox{0.95\linewidth}{
© 2025 IEEE. Personal use of this material is permitted. Permission from IEEE must be obtained for all other uses, in any current or future media, including reprinting/republishing this material for advertising or promotional purposes, creating new collective works, for resale or redistribution to servers or lists, or reuse of any copyrighted component of this work in other works.

This is the author's version of the paper accepted for publication in \textit{IEEE International Symposium on Personal, Indoor and Mobile Radio Communications, 2025}. The final published version is available at IEEE Xplore, \url{https://doi.org/10.1109/PIMRC62392.2025.11274591}.
}}
\end{figure}

\begin{abstract}
Adopting \ac{AI} in \acp{RAN} presents several challenges, limited availability of link-level measurements (e.g., CQI reports), stringent real-time processing constraints (e.g., sub-\qty{1}{ms} per TTI), and network heterogeneity (different spectrum bands, cell types, and vendor equipment). A critical yet often overlooked barrier lies in the computational and memory limitations of \ac{RAN} baseband hardware—particularly in legacy \ac{4G} systems—which typically lack on-chip neural accelerators. As a result, only lightweight \ac{AI} models (under \qty{1}{Mb} and sub-\qty{100}{\mu s} inference time) can be effectively deployed, limiting both their performance and applicability. However, achieving strong generalization across diverse network conditions often requires large-scale models with substantial resource demands. To address this trade-off, this paper investigates \emph{policy distillation} in the context of a reinforcement learning–based link adaptation task. We explore two strategies: \emph{single-policy distillation}, where a scenario-agnostic teacher model is compressed into one generalized student model; and \emph{multi-policy distillation}, where multiple scenario-specific teachers are consolidated into a single generalist student. Experimental evaluations in a high-fidelity, \ac{5G}-compliant simulator demonstrate that both strategies produce compact student models that preserve the teachers' generalization capabilities while complying with the computational and memory limitations of existing \ac{RAN} hardware.
\end{abstract}


\bstctlcite{IEEEexample:BSTcontrol}


\acrodefplural{MDP}[MDPs]{Markov decision processes}
\acrodefplural{RTG}[RTGs]{returns-to-go}

\begin{acronym}
\acro{ACK}{acknowledgement}
\acro{AI}{artificial intelligence}
\acro{ARQ}{automatic repeat request}
\acro{BE}{baseline encoding}
\acro{BC}{behavioral cloning}
  \acro{BCQ}{batch-constrained deep Q-learning}
  \acro{CCTR}{Channel-Conditioned Target Return}
  \acro{CNN}{convolutional neural network}
  \acro{CQL}{conservative Q-learning}
  \acro{CT}{continuous time}
  \acro{CV}{computer vision}
  \acro{DAVG}{discounted average}
  \acro{DLLA}{downlink link adaptation}
  \acro{DP}{dynamic programming}
  \acro{DQN}{deep Q-network}
  \acro{DT}{decision transformer}
  \acro{eMBB}{enhanced mobile broadband}
  \acro{FB}{full buffer}
  \acro{ILLA}{inner-loop link adaptation}
  \acro{gNB}{next generation NodeB}
  \acro{gNB-CU}{gNB centralized unit}
  \acro{gNB-DU}{gNB distributed unit}
  \acro{GPI}{generalized policy iteration}
  \acro{LA}{link adaptation}
  \acro{LSTM}{long short-term memory}
  \acro{LT}{learnable time}
  \acro{KL}{Kullback-Leibler}
  \acro{MBB}{mobile broadband}
  \acro{MC}{Monte Carlo}
  \acro{MDP}{Markov decision process}
  \acro{ML}{machine learning}
  \acro{MLP}{multi-layer perceptron}
  \acro{mMIMO}{massive MIMO}
  \acro{mMTC}{massive machine-type communications}
  \acro{NACK}{negative acknowledgement}
  \acro{NLP}{natural language processing}
  \acro{OLLA}{outer-loop link adaptation}
  \acro{PE}{positional encoding}
  \acro{PHY}{physical layer}
  \acro{RL}{reinforcement learning}
  \acro{RLC}{radio link control}
  \acro{RNN}{recurrent neural network}
  \acro{RTGs}{returns-to-go}
  \acro{RTG}{return-to-go}
  \acro{RvS}{reinforcement learning via supervised learning}
  \acro{SACo}{state-action coverage}
  \acro{SADCo}{state-action density coverage}
  \acro{SCSU}{single-cell single-user}
  \acro{SE}{spectral efficiency}
  \acro{TBS}{transport block size}
  \acro{TD}{temporal difference}
  \acro{TQ}{relative trajectory quality}
  \acro{URLLC}{ultra-reliable low-latency communications}
  \acro{VAE}{variational auto-encoder}
  \acro{2G}{Second Generation}
  \acro{3G}{3$^\text{rd}$~Generation}
  \acro{3GPP}{3$^\text{rd}$~Generation Partnership Project}
  \acro{4G}{4$^\text{th}$~Generation}
  \acro{5G}{5$^\text{th}$~Generation}
  \acro{AA}{Antenna Array}
  \acro{AC}{Admission Control}
  \acro{AD}{Attack-Decay}
  \acro{ADSL}{Asymmetric Digital Subscriber Line}
	\acro{AHW}{Alternate Hop-and-Wait}
  \acro{AMC}{Adaptive Modulation and Coding}
  \acro{AoA}{angle of arrival}
	\acro{AP}{Access Point}
  \acro{APA}{Adaptive Power Allocation}
  \acro{AR}{autoregressive}
  \acro{ARMA}{Autoregressive Moving Average}
  \acro{ATES}{Adaptive Throughput-based Efficiency-Satisfaction Trade-Off}
  \acro{AWGN}{additive white Gaussian noise}
  \acro{BB}{Branch and Bound}
  \acro{BD}{Block Diagonalization}
  \acro{BER}{bit error rate}
  \acro{BF}{Best Fit}
  \acro{BLER}{block error rate}
  \acro{BPC}{Binary power control}
  \acro{BPSK}{Binary Phase-Shift Keying}
  \acro{BPA}{Best \ac{PDPR} Algorithm}
  \acro{BRA}{Balanced Random Allocation}
  \acro{BCRB}{Bayesian Cram\'{e}r-Rao Bound}
  \acro{BS}{base station}
  \acro{CAP}{Combinatorial Allocation Problem}
  \acro{CAPEX}{Capital Expenditure}
  \acro{CBF}{Coordinated Beamforming}
  \acro{CBR}{Constant Bit Rate}
  \acro{CBS}{Class Based Scheduling}
  \acro{CC}{Congestion Control}
  \acro{CDF}{cumulative distribution function}
  \acro{CDMA}{Code-Division Multiple Access}
  \acro{CL}{Closed Loop}
  \acro{CLPC}{Closed Loop Power Control}
  \acro{CNR}{Channel-to-Noise Ratio}
  \acro{CPA}{Cellular Protection Algorithm}
  \acro{CPICH}{Common Pilot Channel}
  \acro{CoMP}{Coordinated Multi-Point}
  \acro{CQI}{channel quality indicator}
  \acro{CRB}{Cram\'{e}r-Rao Bound}
  \acro{CRM}{Constrained Rate Maximization}
	\acro{CRN}{Cognitive Radio Network}
  \acro{CS}{Coordinated Scheduling}
  \acro{CSI}{channel state information}
  \acro{CSIR}{channel state information at the receiver}
  \acro{CSIT}{channel state information at the transmitter}
  \acro{CUE}{cellular user equipment}
  \acro{D2D}{device-to-device}
  \acro{DCA}{Dynamic Channel Allocation}
  \acro{DE}{Differential Evolution}
  \acro{DFT}{Discrete Fourier Transform}
  \acro{DIST}{Distance}
  \acro{DL}{downlink}
  \acro{DMA}{Double Moving Average}
	\acro{DMRS}{demodulation reference signal}
  \acro{D2DM}{D2D Mode}
  \acro{DMS}{D2D Mode Selection}
  \acro{DPC}{Dirty Paper Coding}
  \acro{DRA}{Dynamic Resource Assignment}
  \acro{DSA}{Dynamic Spectrum Access}
  \acro{DSM}{Delay-based Satisfaction Maximization}
  \acro{ECC}{Electronic Communications Committee}
  \acro{EFLC}{Error Feedback Based Load Control}
  \acro{EI}{Efficiency Indicator}
  \acro{eNB}{Evolved Node B}
  \acro{EPA}{Equal Power Allocation}
  \acro{EPC}{Evolved Packet Core}
  \acro{EPS}{Evolved Packet System}
  \acro{ESPRIT}{estimation of signal parameters via rotational invariance}
  \acro{E-UTRAN}{Evolved Universal Terrestrial Radio Access Network}
  \acro{ES}{Exhaustive Search}
  \acro{FDD}{frequency division duplexing}
  \acro{FDM}{Frequency Division Multiplexing}
  \acro{FER}{Frame Erasure Rate}
  \acro{FF}{Fast Fading}
  \acro{FIM}{Fisher information matrix}
  \acro{FSB}{Fixed Switched Beamforming}
  \acro{FST}{Fixed SNR Target}
  \acro{FTP}{File Transfer Protocol}
  \acro{GA}{Genetic Algorithm}
  \acro{GBR}{Guaranteed Bit Rate}
  \acro{GLR}{Gain to Leakage Ratio}
  \acro{GOS}{Generated Orthogonal Sequence}
  \acro{GPL}{GNU General Public License}
  \acro{GRP}{Grouping}
  \acro{HARQ}{hybrid automatic repeat request}
  \acro{HMS}{Harmonic Mode Selection}
  \acro{HOL}{Head Of Line}
  \acro{HSDPA}{High-Speed Downlink Packet Access}
  \acro{HSPA}{High Speed Packet Access}
  \acro{HTTP}{HyperText Transfer Protocol}
  \acro{ICMP}{Internet Control Message Protocol}
  \acro{ICI}{Intercell Interference}
  \acro{ID}{Identification}
  \acro{ISAC}{integrated sensing and communication}
  \acro{IEEE}{Institute of Electrical and Electronics Engineers}
  \acro{IETF}{Internet Engineering Task Force}
  \acro{ILP}{Integer Linear Program}
  \acro{JRAPAP}{Joint RB Assignment and Power Allocation Problem}
  \acro{UID}{Unique Identification}
  \acro{IID}{Independent and Identically Distributed}
  \acro{IIR}{Infinite Impulse Response}
  \acro{ILP}{Integer Linear Problem}
  \acro{IMT}{International Mobile Telecommunications}
  \acro{INV}{Inverted Norm-based Grouping}
  \acro{IoT}{Internet of Things}
  \acro{IP}{Internet Protocol}
  \acro{IPv6}{Internet Protocol Version 6}
  \acro{ISD}{Inter-Site Distance}
  \acro{ISI}{Inter Symbol Interference}
  \acro{ITU}{International Telecommunication Union}
  \acro{JOAS}{Joint Opportunistic Assignment and Scheduling}
  \acro{JOS}{Joint Opportunistic Scheduling}
  \acro{JP}{Joint Processing}
  \acro{JS}{Jump-Stay}
  \acro{KKT}{Karush-Kuhn-Tucker}
  \acro{L3}{Layer-3}
  \acro{LAC}{Link Admission Control}
  \acro{LC}{Load Control}
  \acro{LOS}{Line of Sight}
  \acro{LP}{Linear Programming}
  \acro{LS}{least squares}
  \acro{LTE}{Long Term Evolution}
  \acro{LTE-A}{LTE-Advanced}
  \acro{LTE-Advanced}{Long Term Evolution Advanced}
  \acro{M2M}{Machine-to-Machine}
  \acro{MAC}{Medium Access Control}
  \acro{MANET}{Mobile Ad hoc Network}
  \acro{MCS}{modulation and coding scheme}
  \acro{MDB}{Measured Delay Based}
  \acro{MDI}{Minimum D2D Interference}
  \acro{MF}{Matched Filter}
  \acro{MG}{Maximum Gain}
  \acro{MH}{Multi-Hop}
  \acro{MIMO}{multiple input multiple output}
  \acro{MINLP}{Mixed Integer Nonlinear Programming}
  \acro{MIP}{Mixed Integer Programming}
  \acro{MISO}{Multiple Input Single Output}
  \acro{MLE}{maximum likelihood estimator}
  \acro{MLWDF}{Modified Largest Weighted Delay First}
  \acro{MME}{Mobility Management Entity}
  \acro{MMSE}{minimum mean squared error}
  \acro{MOS}{Mean Opinion Score}
  \acro{MPF}{Multicarrier Proportional Fair}
  \acro{MRA}{Maximum Rate Allocation}
  \acro{MR}{Maximum Rate}
  \acro{MRC}{Maximum Ratio Combining}
  \acro{MRT}{Maximum Ratio Transmission}
  \acro{MRUS}{Maximum Rate with User Satisfaction}
  \acro{MS}{mobile station}
  \acro{MSE}{mean squared error}
  \acro{MSI}{Multi-Stream Interference}
  \acro{MTC}{Machine-Type Communication}
  \acro{MTSI}{Multimedia Telephony Services over IMS}
  \acro{MTSM}{Modified Throughput-based Satisfaction Maximization}
  \acro{MU-MIMO}{multiuser multiple input multiple output}
  \acro{MU}{multi-user}
  \acro{MUSIC}{multiple signal classification}
  \acro{NAS}{Non-Access Stratum}
  \acro{NB}{Node B}
  \acro{NE}{Nash equilibrium}
  \acro{NCL}{Neighbor Cell List}
  \acro{NLOS}{Non-Line of Sight}
  \acro{NMSE}{Normalized Mean Square Error}
  \acro{NORM}{Normalized Projection-based Grouping}
  \acro{NP}{Non-Polynomial Time}
  \acro{NR}{New Radio}
  \acro{NRT}{Non-Real Time}
  \acro{NSPS}{National Security and Public Safety Services}
  \acro{O2I}{Outdoor to Indoor}
  \acro{OFDMA}{orthogonal frequency division multiple access}
  \acro{OFDM}{orthogonal frequency division multiplexing}
  \acro{OFPC}{Open Loop with Fractional Path Loss Compensation}
	\acro{O2I}{Outdoor-to-Indoor}
  \acro{OL}{Open Loop}
  \acro{OLPC}{Open-Loop Power Control}
  \acro{OL-PC}{Open-Loop Power Control}
  \acro{OPEX}{Operational Expenditure}
  \acro{ORB}{Orthogonal Random Beamforming}
  \acro{JO-PF}{Joint Opportunistic Proportional Fair}
  \acro{OSI}{Open Systems Interconnection}
  \acro{PAIR}{D2D Pair Gain-based Grouping}
  \acro{PAPR}{Peak-to-Average Power Ratio}
  \acro{P2P}{Peer-to-Peer}
  \acro{PC}{Power Control}
  \acro{PCI}{Physical Cell ID}
  \acro{PDF}{probability density function}
  \acro{PDPR}{pilot-to-data power ratio}
  \acro{PER}{Packet Error Rate}
  \acro{PF}{Proportional Fair}
  \acro{P-GW}{Packet Data Network Gateway}
  \acro{PL}{Pathloss}
  \acro{PPR}{pilot power ratio}
  \acro{PRB}{physical resource block}
  \acro{PROJ}{Projection-based Grouping}
  \acro{ProSe}{Proximity Services}
  \acro{PS}{Packet Scheduling}
  \acro{PSAM}{pilot symbol assisted modulation}
  \acro{PSO}{Particle Swarm Optimization}
  \acro{PZF}{Projected Zero-Forcing}
  \acro{QAM}{Quadrature Amplitude Modulation}
  \acro{QoS}{Quality of Service}
  \acro{QPSK}{Quadri-Phase Shift Keying}
  \acro{RAISES}{Reallocation-based Assignment for Improved Spectral Efficiency and Satisfaction}
  \acro{RAN}{radio access network}
  \acro{RAT}{Radio Access Technology}
  \acro{RATE}{Rate-based}
  \acro{RB}{resource block}
  \acro{RBG}{Resource block broup}
  \acro{REF}{Reference Grouping}
  \acro{RM}{Rate Maximization}
  \acro{RNC}{Radio Network Controller}
  \acro{RND}{Random Grouping}
  \acro{RRA}{Radio Resource Allocation}
  \acro{RRM}{radio resource management}
  \acro{RSCP}{Received Signal Code Power}
  \acro{RSRP}{Reference Signal Receive Power}
  \acro{RSRQ}{Reference Signal Receive Quality}
  \acro{RR}{Round Robin}
  \acro{RRC}{Radio Resource Control}
  \acro{RSSI}{Received Signal Strength Indicator}
  \acro{RT}{Real Time}
  \acro{RU}{Resource Unit}
  \acro{RUNE}{RUdimentary Network Emulator}
  \acro{RV}{Random Variable}
  \acro{SAC}{Session Admission Control}
  \acro{SCM}{Spatial Channel Model}
  \acro{SC-FDMA}{Single Carrier - Frequency Division Multiple Access}
  \acro{SD}{Soft Dropping}
  \acro{S-D}{Source-Destination}
  \acro{SDPC}{Soft Dropping Power Control}
  \acro{SDMA}{Space-Division Multiple Access}
  \acro{SER}{Symbol Error Rate}
  \acro{SES}{Simple Exponential Smoothing}
  \acro{S-GW}{Serving Gateway}
  \acro{SINR}{signal-to-interference-plus-noise ratio}
  \acro{SI}{Satisfaction Indicator}
  \acro{SIP}{Session Initiation Protocol}
  \acro{SISO}{single input single output}
  \acro{SIMO}{Single Input Multiple Output}
  \acro{SIR}{signal-to-interference ratio}
  \acro{SLNR}{Signal-to-Leakage-plus-Noise Ratio}
  \acro{SMA}{Simple Moving Average}
  \acro{SNR}{signal-to-noise ratio}
  \acro{SORA}{Satisfaction Oriented Resource Allocation}
  \acro{SORA-NRT}{Satisfaction-Oriented Resource Allocation for Non-Real Time Services}
  \acro{SORA-RT}{Satisfaction-Oriented Resource Allocation for Real Time Services}
  \acro{SPF}{Single-Carrier Proportional Fair}
  \acro{SRA}{Sequential Removal Algorithm}
  \acro{SRS}{Sounding Reference Signal}
  \acro{SSB}{synchronisation signal block}
  \acro{SU-MIMO}{single-user multiple input multiple output}
  \acro{SU}{Single-User}
  \acro{SVD}{Singular Value Decomposition}
  \acro{TCP}{Transmission Control Protocol}
  \acro{TDD}{time division duplexing}
  \acro{TDMA}{Time Division Multiple Access}
  \acro{TETRA}{Terrestrial Trunked Radio}
  \acro{TP}{Transmit Power}
  \acro{TPC}{Transmit Power Control}
  \acro{TTI}{transmission time interval}
  \acro{TTR}{Time-To-Rendezvous}
  \acro{TSM}{Throughput-based Satisfaction Maximization}
  \acro{TU}{Typical Urban}
  \acro{UE}{user equipment}
  \acro{UEPS}{Urgency and Efficiency-based Packet Scheduling}
  \acro{UL}{uplink}
  \acro{UMTS}{Universal Mobile Telecommunications System}
  \acro{URI}{Uniform Resource Identifier}
  \acro{URM}{Unconstrained Rate Maximization}
  \acro{UT}{user terminal}
  \acro{VR}{Virtual Resource}
  \acro{VoIP}{Voice over IP}
  \acro{WAN}{Wireless Access Network}
  \acro{WCDMA}{Wideband Code Division Multiple Access}
  \acro{WF}{Water-filling}
  \acro{WiMAX}{Worldwide Interoperability for Microwave Access}
  \acro{WINNER}{Wireless World Initiative New Radio}
  \acro{WLAN}{Wireless Local Area Network}
  \acro{WMPF}{Weighted Multicarrier Proportional Fair}
  \acro{WPF}{Weighted Proportional Fair}
  \acro{WSN}{Wireless Sensor Network}
  \acro{WWW}{World Wide Web}
  \acro{XIXO}{(Single or Multiple) Input (Single or Multiple) Output}
  \acro{ZF}{zero-forcing}
  \acro{ZMCSCG}{Zero Mean Circularly Symmetric Complex Gaussian}
\end{acronym}

\acresetall 

\section{Introduction}\label{sec:1}

\Ac{AI} is increasingly recognized as a key enabler of improved performance and efficiency in \acp{RAN}. \ac{AI}-driven solutions have been explored across a broad range of \ac{RAN} functions, including channel estimation and symbol decoding at the physical layer~\cite{SPM+:19, SoD:22}; fast control loops, such as link adaptation and radio resource management, at the medium access layer~\cite{KHV+:22, Chen:23, calabrese:18, SKC:22}; and slower functionalities such as mobility management, load balancing, and network configuration~\cite{AoF:23, GSE+:22}. Factors such as user mobility, interference, traffic fluctuations, and the heterogeneity of both \ac{RAN} infrastructure and \ac{UE} introduce significant variability. As a result, achieving reliable performance in real-world \ac{RAN} deployments requires \ac{AI} models that can adapt to diverse and dynamic network conditions~\cite{generalization_2024}. In other words, models must generalize effectively and perform robustly, even on previously unseen data~\cite{tu2020understanding}.

Training an \ac{AI} model for a \ac{RAN} function to generalize across diverse \ac{RAN} environments—rather than fine-tuning separate models for specific conditions, such as those encountered in a single radio cell—is essential for scalable \ac{AI} integration in \acp{RAN}~\cite{generalization_2024}. Reducing model proliferation enables more efficient and cost-effective lifecycle management. In contrast, models fine-tuned to specific network conditions often lack the robustness required to adapt to changes in network and environmental dynamics, necessitating frequent updates and compromising performance consistency. However, achieving effective model generalization requires extensive and diverse training data, as well as sufficiently large models capable of learning and extrapolating complex patterns from large datasets~\cite{tu2020understanding}.

Although \acp{RAN} are data-rich systems, a fundamental challenge in deploying \ac{AI} within existing infrastructure is the computational constraint of legacy baseband hardware. Designed primarily for efficient signal processing and protocol execution, \ac{4G} and \ac{5G} baseband units were not dimensioned to support large \ac{AI} models. As a result, a mismatch often arises between the demands of \ac{AI}-based network optimization and the limited hardware resources available for inference in \ac{RAN} systems. This restricts the size and complexity of models that can be practically deployed, thereby limiting their ability to generalize effectively.

Knowledge distillation~\cite{gou2021knowledge} has been proposed to address the challenge of deploying complex models on resource-constrained devices. It enables knowledge transfer from a large, high-performing teacher model to a smaller, computationally efficient student model, with negligible degradation in accuracy~\cite{hinton2015distilling}. Originally introduced for supervised learning tasks~\cite{BCR:06, hinton2015distilling}, recent work~\cite{RGG+:15, SoF:19} has extended these techniques to \emph{policy distillation} in reinforcement learning for control applications. However, the ability of policy distillation to retain the generalization capability of the teacher model—and its applicability to practical problems in \acp{RAN}—remains relatively unexplored.

This paper explores policy distillation as a method to replace traditional \ac{RAN} functions with \ac{AI}-based alternatives under realistic \ac{RAN} operational constraints. Our goal is to distill \ac{AI} models that comply with \ac{RAN} baseband hardware constraints while retaining the sophisticated decision-making capabilities of large, computationally intensive teacher models. Focusing on the \ac{LA} function, we employ \emph{domain randomization} to generate diverse training datasets from varied simulated environments. These datasets are then used to train compact, computationally efficient student models that closely replicate the behavior of high-capacity teacher models. We evaluate two distillation strategies: \emph{single-policy} distillation, which compresses a single scenario-agnostic teacher into a compact student model; and \emph{multi-policy} distillation, which integrates knowledge from multiple scenario-specific teachers into a generalist student model. Experimental evaluations in a \ac{5G}-compliant simulator confirm both strategies dramatically reduce model size while preserving strong generalization.

\section{The Hidden Challenges of AI in RAN}

Significant research has been devoted to AI-driven solutions for various \ac{RAN} functions, examining algorithms, models, training methods, and data collection strategies. However, limited attention has been paid to the \emph{real-time processing constraints} that hinder the deployment of AI in practical \ac{RAN} settings. Existing \ac{RAN} baseband hardware—typically based on FPGA, ASIC, or DSP architectures optimized for specialized, repetitive signal processing tasks (e.g., FFT, modulation/demodulation) and protocol execution—was not originally designed to support AI workloads, such as tensor computations or floating-point operations. Consequently, current hardware lacks the flexibility and computational headroom (e.g., memory and processing resources) required to effectively support large AI models. Additionally, baseband units commonly operate at full computational capacity under strict latency and power constraints, necessitating minimal—ideally zero—additional overhead from any AI integration.

Another crucial constraint is the stringent latency requirement of critical user-plane functions in Layer 1 (L1) and Layer 2 (L2), such as resource scheduling and link adaptation. These functions must execute within tens of microseconds—for example, link adaptation (LA) typically requires less than ten microseconds—thereby significantly limiting permissible inference latency. These latency constraints restrict AI model complexity, memory footprint, and per-inference computational cost, often limiting feasible AI models to sizes no larger than a few tens of kilobytes due to limited on-chip memory and strict inference-time budgets. Collectively, these constraints underscore the need for efficient AI deployment strategies tailored specifically to legacy \ac{RAN} infrastructure.

To address these challenges, we investigate policy distillation—a technique that transfers knowledge from large, computationally intensive AI models to smaller models suitable for resource-constrained hardware. We apply this approach to LA, an L2-critical-loop \ac{RAN} function executed multiple times (once per scheduled UE) on a sub-millisecond timescale.

\section{Problem formulation}\label{sec:2}

We consider the problem of \ac{LA} in the downlink of a \ac{RAN}, where the transmission rate is dynamically adjusted to match the link capacity by optimizing the \ac{MCS} parameters based on time- and frequency-varying channel conditions and radio interference. In \ac{5G} systems, the \ac{MCS} parameters—comprising the modulation order and code rate—are represented by an \ac{MCS} index~\cite{3rd_generation_partnership_project_3gpp_technical_2024}, denoted by integer values $m$ ranging from $0$ to $M-1$.

LA algorithms deployed in present-day \acp{RAN} typically rely on two control loops: an inner loop that maps the latest \ac{CSI} reported by a \ac{UE} to an initial \ac{SINR} estimate, and an outer loop that adjusts this estimate based on \ac{HARQ} feedback to maximize spectral efficiency while meeting a target \ac{BLER}. However, the outer loop converges slowly~\cite{BGA:16} to a fixed \ac{BLER} target; therefore, it struggles to adapt to dynamic channel conditions under bursty traffic. In contrast, \ac{RL}-based approach can learn a policy that predicts the optimal \ac{MCS} index in real time, offering the potential to overcome the limitations of rule-based \ac{LA} algorithms.

\subsection{RL-based Link Adaptation}

\begin{figure}[t]
\centering
\includegraphics[width=1\columnwidth]{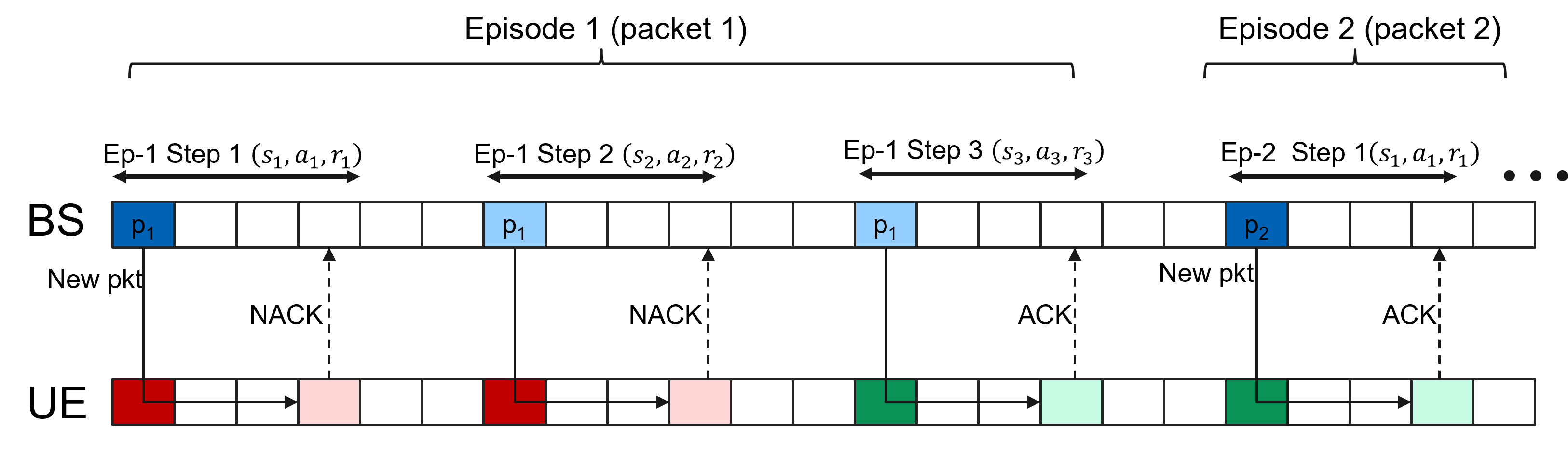}
\caption{Example of \ac{MDP} episode definition for \ac{LA}.}
\label{fig:episode}
\end{figure}

We model the \ac{LA} problem as an episodic \ac{MDP} $\mathcal{M} = \left\langle\mathcal{S}, \mathcal{A}, \mathcal{P}, \mathcal{R}, \gamma \right\rangle$, where each episode corresponds to the lifespan of a \ac{UE} data packet—from its first transmission to either successful reception or packet drop after $N$ transmission attempts, as illustrated in~\cref{fig:episode}. The action space $\mathcal{A} = \{a \mid a \in \{0,\dots,M-1\}\}$ comprises $M$ discrete values, representing the \ac{MCS} indices used to adapt the link spectral efficiency. In our evaluations, we consider 28 \ac{MCS} indices (i.e., $M = 28$), as defined in Table 5.1.3.1-2 of the \ac{5G} specifications~\cite{3rd_generation_partnership_project_3gpp_technical_2024}. 

Following~\cite{DWT+:25}, we use a reward function $r_{n} = r(s_{n}, a_{n}) \in \mathcal{R}$ for each individual packet transmission, defined as
\begin{equation}
r_{n} =
\begin{cases} 
    SE_n, & \text{if successfully received at } n^{\mathrm{th}} \text{ TX},\\
    -\alpha \cdot n, & \text{if unsuccessfully received at } n^{\mathrm{th}} \text{ TX},
\end{cases}
\label{reward}
\end{equation}
where $n \in \{0, \dots, N-1\}$ is the packet transmission index, $\alpha \in \mathbb{R}_{+}$ is a penalty coefficient for retransmissions, and $SE_n$ denotes the spectral efficiency achieved upon successful reception in the $n^{\mathrm{th}}$ transmission attempt.

To support model generalization across the \ac{RAN} environment during training, we consider two key enablers. First, we design the state $s \in \mathcal{S}$ to encapsulate a mixture of semi-static and dynamic information. The semi-static information serves two purposes: (i) to characterize the heterogeneity of the \ac{RAN} deployment and its configuration relative to the \ac{UE}—including deployment type, location, orientation, interrelations among network sites or radio cells, and parameters such as antenna array type, carrier frequency, bandwidth, and transmit power—and (ii) to describe the \ac{UE}'s characteristics (e.g., chipset and capabilities), enabling the model to operate across a diverse set of devices. The dynamic information, in contrast, captures state variables relevant to \ac{LA} behavior, such as path loss, \ac{CSI}, \ac{HARQ} feedback, and many more; see, e.g.,~\cite{DWT+:25} for the complete list of states.

Additionally, we leverage a distributed training architecture combined with domain randomization to improve model robustness by exposing the agent to varied and diverse environmental conditions. Domain randomization introduces controlled stochasticity into the training environment to bridge the simulation-to-reality gap, as further described in~\cref{sec:6}.

\section{Policy Distillation}\label{sec:3}

To address the challenge of achieving model generalization while adhering to the constraints imposed by \ac{RAN} hardware, we merge knowledge distillation with domain randomization.

Knowledge distillation, referred to as \emph{policy distillation} in the \ac{RL} context~\cite{RGG+:15}, involves transferring a policy learned by an agent (the teacher), through interaction with an environment, into a smaller, simpler model (the student) without significant performance degradation. Policy distillation can be performed in both online and offline settings~\cite{RGG+:15, rusu2016policydistillation, SoF:19}. In \emph{online policy distillation}, the student model learns concurrently with the teacher, as the latter improves its policy through ongoing interaction with the environment. In contrast, \emph{offline policy distillation} assumes that the teacher has been fully trained. The student is then trained using a dataset generated by the teacher’s fixed policy.

In this paper, we only focus on the offline approach, assuming a trained \ac{DQN} policy for \ac{LA} as the teacher, following the design presented in~\cref{sec:2}.

\subsection{Offline Policy Distillation}

Following~\cite{rusu2016policydistillation}, we consider a distillation dataset $\mathcal{D}^T = \{(s_i, \mathbf{q}_{i}^{T})\}_{i}$, where each sample $i$ consists of a state $s_{i} \in \mathcal{S}$ from the \ac{LA} function and a vector of unnormalized Q-values $\mathbf{q}_{i}^{T} = \pi^{T}(s_{i})$ produced by the teacher policy $\pi^{T}$. These Q-values serve as target outputs for training a student policy $\pi^{S}$.

To guide the training, we employ the \ac{KL} divergence loss, defined as
\begin{equation}
L_{\mathrm{KL}}(\mathcal{D}^T, \pi^S) = \sum_{i=1}^{\mid\mathcal{D}^T\mid} \mathrm{softmax}\left(\frac{\mathbf{q}^T_i}{\tau}\right) \ln \frac{\mathrm{softmax}(\frac{\mathbf{q}^T_i}{\tau})}{\mathrm{softmax}(\mathbf{q}^S_i)} \;.
\label{KL}
\end{equation}
Here, a softmax transformation is applied to both the teacher’s Q-values $\mathbf{q}_{i}^{T}$, scaled by a temperature parameter $\tau > 0$, and the student’s Q-values $\mathbf{q}_{i}^{S}$, which are left unscaled. In the context of policy distillation, Rusu~\textit{et al.}~\cite{rusu2016policydistillation} recommend using lower values of $\tau$ to sharpen the output distribution. Since Q-values represent the expected future discounted rewards for each possible action, sharpening accentuates the differences in action preferences and improves the fidelity of the transferred~policy.

Unlike prior work, our objective is to distill a student policy $\pi^{S}$ that retains the teacher’s generalization capability while being significantly more compact. To this end, we revisit and extend two offline policy distillation strategies: single- and multi-policy distillation~\cite{rusu2016policydistillation}.

\subsection{Generalization via Single-Policy Distillation}

Our first approach combines single-policy distillation with domain randomization. In this case, we distill a student policy $\pi^S$ from the knowledge of a single expert-level teacher~$\pi^T$. However, to retain the teacher's ability to generalize across the \ac{RAN} environment, the distillation dataset $\mathcal{D}^T$ must capture the teacher’s behavior under diverse radio communication conditions. To this end, we also apply domain randomization when generating the distillation samples $(s_i, \mathbf{q}_i^T)$, as described in~\cref{sec:6}, ensuring that the state samples $s_i$ are drawn from heterogeneous network deployments with varying configurations, radio environments, traffic and interference types, load conditions, user device types, and so on.

Practical \ac{RAN} deployments are, by nature, a source of extreme diversity due to the variety of radio environments, network configurations, traffic patterns, and interference distributions encountered across radio cells. Therefore, if a teacher model is trained using the collective experience of distributed actors operating in different radio cells, a diverse distillation dataset $\mathcal{D}^T$ can be constructed by reusing data stored in the teacher’s replay memory. Specifically, the trained teacher can be evaluated on state observations $s_i$ already present in the replay memory used during training.

\subsection{Generalization via Multi-Policy Distillation}

Our second approach is inspired by multi-policy distillation. Training a large \ac{RL} teacher model from the collective experience of radio cells across a live \ac{RAN} can be costly and may negatively impact \ac{RAN} performance due to the exploration process. As an alternative path to achieving model generalization under hardware constraints, we propose training multiple teacher models using less invasive methods and applying multi-policy distillation to induce generalization by fusing their policies into a single student model.

Specifically, we consider $N$ independently trained teacher policies $\pi^{T_j}$, each specialized—rather than generalized—for a specific network environment or deployment. This allows each teacher model to be dimensioned according to the hardware constraints of the target \ac{RAN}. For instance, different teacher models may be trained separately using drive tests in specific network contexts (e.g., urban, rural, high-speed), as well as with proprietary networks, lab testbeds, or high-fidelity network simulators.

After training a set of teacher models, we generate the corresponding distillation datasets, denoted as $\mathcal{D}^{T_j} = \{(s_i, \mathbf{q}_i^{T_j})\}_{i=0}^{D_j}$ for all $j \in\{0, \dots, N-1\}$, following the same methodology as in single-policy distillation. We then shuffle and aggregate these datasets to train a unified student model. Since the student learns from the combined behavior of multiple specialized teachers, the resulting policy generalizes effectively across diverse \ac{RAN} environments.

\section{Numerical examples}\label{sec:6}

To evaluate the potential of knowledge distillation for facilitating the deployment of \ac{AI} in \acp{RAN}, we focus on its ability to reduce model size to meet the hardware limitations of current \ac{RAN} technology while achieving the performance and retaining the generalization capability of a larger teacher model. To this end, we consider both single-policy and multi-policy distillation applied to a \ac{RL} policy for \ac{MCS} index selection in downlink \ac{LA}, and compare the performance of a larger teacher model against multiple student models of varying size across a range of performance metrics. To assess whether the student models retain the generalization ability of the teacher, we evaluate them using three representative benchmark scenarios not encountered during training. Furthermore, we demonstrate that training a small model directly with \ac{RL}—rather than applying policy distillation—fails to achieve comparable performance and generalization.

\subsection{Training Setup}

To train the teacher models, similar to~\cite{DWT+:25}, we use a large-scale distributed training architecture, in which a single learner updates the model using the collective experience of multiple distributed actors and broadcasts the updated model parameters to the actors. Each actor interacts with several parallel simulations based on a 5G-compliant event-driven network simulator. Each training simulation emulates a time-division duplex system operating at a \qty{3.5}{GHz} carrier frequency, with physical layer numerology $\mu = 0$, and \ac{SU-MIMO} transmission according to a parameter configuration sampled from \cref{tab:ran_params}. However, we adopt a different training approach for teachers used in single-policy and multi-policy distillation, as discussed later.

\begin{table}[ht!]
\centering
\renewcommand{\arraystretch}{1.2}
\begin{tabular}{ll}
\hline
\textbf{Parameter}           & \textbf{Value range}                                     \\
\hline
Antenna array                & $1\times 2\times 2$ MIMO (4)                             \\  
                             & $8\times 4\times 2$ mMIMO (64)                           \\  
Cell radius                  & \{166, 300, 600, 900, 1200\} \unit{m}                    \\  
Bandwidth                    & \{20, 40, 50, 80, 100\} \unit{MHz}                       \\  
Number of sub-bands          & \{51, 106, 133, 217, 273\}                               \\  
DL TX power                  & \{20, 40, 50, 80, 100\} \unit{W}                         \\  
UE antennas                  & \{2, 4\}                                                 \\  
Maximum TX rank              & \{2, 4\}                                                 \\  
Maximum DL TX                & 5                                                        \\  
Number FB UEs                & \{1, 5, 10\}                                             \\  
Number MBB UEs               & \{5, 10, 25, 50, 100, 150\}                              \\  
Speed UE FB                  & \{0.67, 10, 15, 30\} \unit{m/s}                          \\  
Speed UE MBB                 & \{0.67, 1.5, 3\} \unit{m/s}                              \\  
Indoor probability           & \{0.2, 0.4, 0.8\}                                        \\  
\hline
\end{tabular}
\caption{Parameters for domain randomization.}
\label{tab:ran_params}
\end{table}

For both single- and multi-policy distillation, we employ a teacher architecture consisting of a 7-layer \ac{MLP} with 128 neurons per layer ($7\times128$), comprising approximately \qty{105}{k} parameters. We distill three student models of different sizes: (i) a 4-layer \ac{MLP} with 64 neurons per layer ($4\times64$); (ii) a 4-layer \ac{MLP} with 32 neurons per layer ($4\times32$); and (iii) a 3-layer \ac{MLP} with 32 neurons per layer ($3\times32$). These correspond to approximately \qty{15}{k}, \qty{5}{k}, and \qty{3.5}{k} model parameters, respectively.

\subsection{Testing Setup}

To assess model generalization, we evaluate all models against three benchmark scenarios not explicitly encountered during training: \ac{SCSU}, \ac{MIMO}, and \ac{mMIMO}, each assuming three cells with \ac{FB} traffic and 10 \acp{UE}.

We begin by evaluating the teacher model against the \ac{LA} baseline commonly adopted in present-day \acp{RAN}, as described in~\cref{sec:3}. In our benchmark scenarios, the teacher achieves an average \ac{UE} throughput gain of $12\%$ and a \ac{SE} improvement of $8\%$ over the baseline, demonstrating its ability to perform well in diverse scenarios not seen during training. However, since our primary objective is to assess whether policy distillation can retain the teacher’s generalization capability while significantly reducing model size, we focus our analysis on comparing student performance with that of the corresponding teacher across the three benchmark scenarios.

We consider three performance metrics: average \ac{UE} throughput ($T$), \ac{BLER}, and episodic reward ($r$). The results are reported using the \textit{relative gain/loss}. Values close to zero indicate that the student models closely approximate the teacher’s behavior. For example, the relative throughput gain/loss ($\Delta T$) is computed as:
$$
\Delta T = 100 \times \frac{T_{\mathrm{student}} - T_{\mathrm{teacher}}}{T_{\mathrm{teacher}}} \;.
$$
Analogous definitions apply to $\Delta \mathrm{\ac{BLER}}$ and $\Delta r$. Finally, to assess whether the distillation process effectively transfers the teacher’s policy, we compare the \ac{PDF} of the actions returned by the teacher and the corresponding student models in each benchmark scenario.

\begin{table*}[!h]
\centering
\renewcommand{\arraystretch}{1.3}
\begin{tabular}{l|c|ccc|ccc|ccc}
\hline
\textbf{Distillation} & \textbf{Student Model} & \multicolumn{3}{c|}{\textbf{MIMO Scenario}} & \multicolumn{3}{c|}{\textbf{mMIMO Scenario}} & \multicolumn{3}{c}{\textbf{SCSU Scenario}} \\
\cline{3-11}
 & & $\Delta T$ & $\Delta \mathrm{BLER}$ & $\Delta r$ & $\Delta T$ & $\Delta \mathrm{BLER}$ & $\Delta r$ & $\Delta T$ & $\Delta \mathrm{BLER}$ & $\Delta r$ \\
\hline
\multirow{3}{*}{Single-policy}
& Student $4\times 64$ & $+0.04\%$ & $+4.7\%$ & $-0.5\%$ & $-0.02\%$ & $+2.4\%$ & $-0.8\%$ & $-0.21\%$ & $+2.1\%$ & $-0.8\%$ \\
& Student $4 \times 32$ & $+0.14\%$ & $+2.3\%$ & $-0.9\%$ & $+0.15\%$ & $0.0\%$ & $-1.2\%$ & $+0.02\%$ & $-6.6\%$ & $-1.1\%$ \\
& Student $3 \times 32$ & $-0.17\%$ & $+6.5\%$ & $-1.7\%$ & $+0.09\%$ & $+4.7\%$ & $-1.7\%$ & $-0.54\%$ & $+6.5\%$ & $-2.2\%$ \\
\hline
\multirow{3}{*}{Multi-policy}
& Student $4 \times 64$ & $-1.2\%$ & $+4.8\%$ & $-2.5\%$ & $-1.0\%$ & $-2.7\%$ & $-0.8\%$ & $-1.6\%$ & $+2.5\%$ & $-0.8\%$ \\
& Student $4 \times 32$ & $-2.3\%$ & $+2.9\%$ & $-3.3\%$ & $-1.9\%$ & $-3.7\%$ & $-1.2\%$ & $-1.9\%$ & $-2.5\%$ & $-1.6\%$ \\
& Student $3 \times 32$ & $-2.8\%$ & $+5.7\%$ & $-3.7\%$ & $-2.3\%$ & $-4.0\%$ & $-1.5\%$ & $-2.2\%$ & $0.0\%$ & $-1.9\%$ \\
\hline
\end{tabular}
\caption{Relative gain/loss (\%) of student models with respect to their teacher models, across performance metrics and benchmark scenarios.}
\label{tab:student_performance}
\end{table*}

\subsection{Generalization via Single-Policy Distillation}

We apply single-policy distillation to an \ac{RL} policy for \ac{MCS} index selection in downlink \ac{LA}, trained to generalize across the \ac{RAN} environment. To this end, we combine the previously introduced distributed \ac{RL} training architecture with domain randomization to improve model resilience to uncertainties in the \ac{RAN} environment, such as network deployment and configuration, traffic conditions, and user population.

Specifically, we train the teacher model using 5000 simulations, distributed across 16 actors, each with randomized network parameters based on \cref{tab:ran_params}. Each simulation lasts 3 seconds and consists of one to three tri-sectorial radio sites, randomly configured as \ac{MIMO} or \ac{mMIMO}, according to antenna characteristics in Table~\ref{tab:ran_params}. Site parameters such as inter-site distance, cell radius, bandwidth, and downlink transmit power are also randomly sampled from Table~\ref{tab:ran_params}. Further randomization includes cell loads, traffic types, indoor/outdoor \ac{UE} ratios, and \ac{UE} receiver types. \acp{UE} are generated with random counts of \ac{FB} and \ac{MBB} traffic, with traffic patterns modeled from real-world data. \ac{UE} configurations—including the number of antenna elements, speed, and receiver type—are randomized to reflect hardware and algorithmic differences across devices (e.g., \ac{CSI} estimation).

To enable the distillation process to retain the generalization capability of the teacher model, we further apply domain randomization to generate the distillation dataset $\mathcal{D}^T$ by testing the teacher against 1000 randomized network simulations, resulting in approximately 20 million samples—providing sufficient data diversity for effective distillation.

\begin{figure*}[htbp]
    \centering
    \begin{subfigure}{0.3\textwidth}
        \centering
        \includegraphics[width=\linewidth]{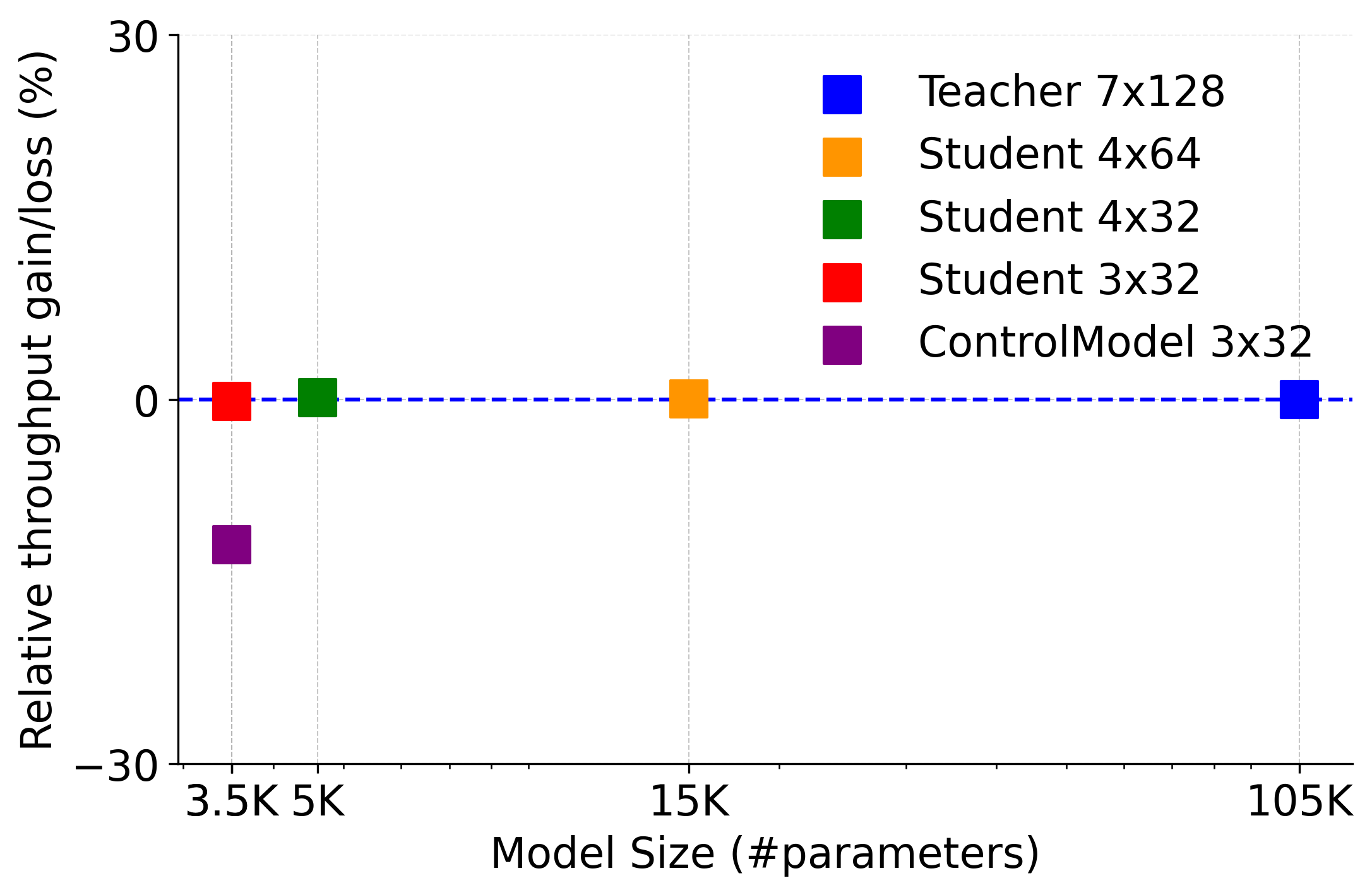}
        \caption{MIMO}
        \label{d}
    \end{subfigure}
    \hfill
    \begin{subfigure}{0.3\textwidth}
        \centering
        \includegraphics[width=\linewidth]{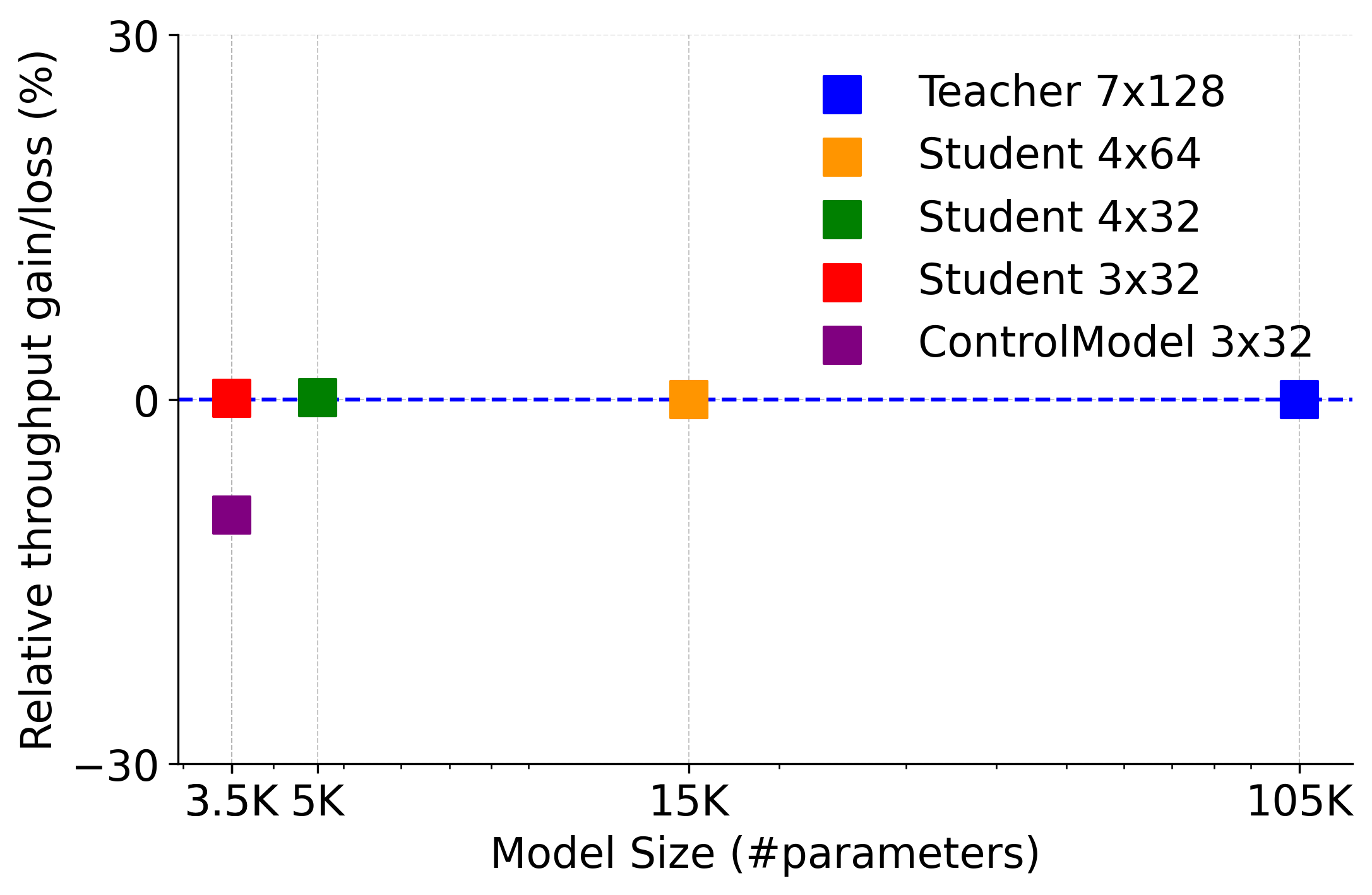}
        \caption{mMIMO}
        \label{e}
    \end{subfigure}
    \hfill
    \begin{subfigure}{0.3\textwidth}
        \centering
        \includegraphics[width=\linewidth]{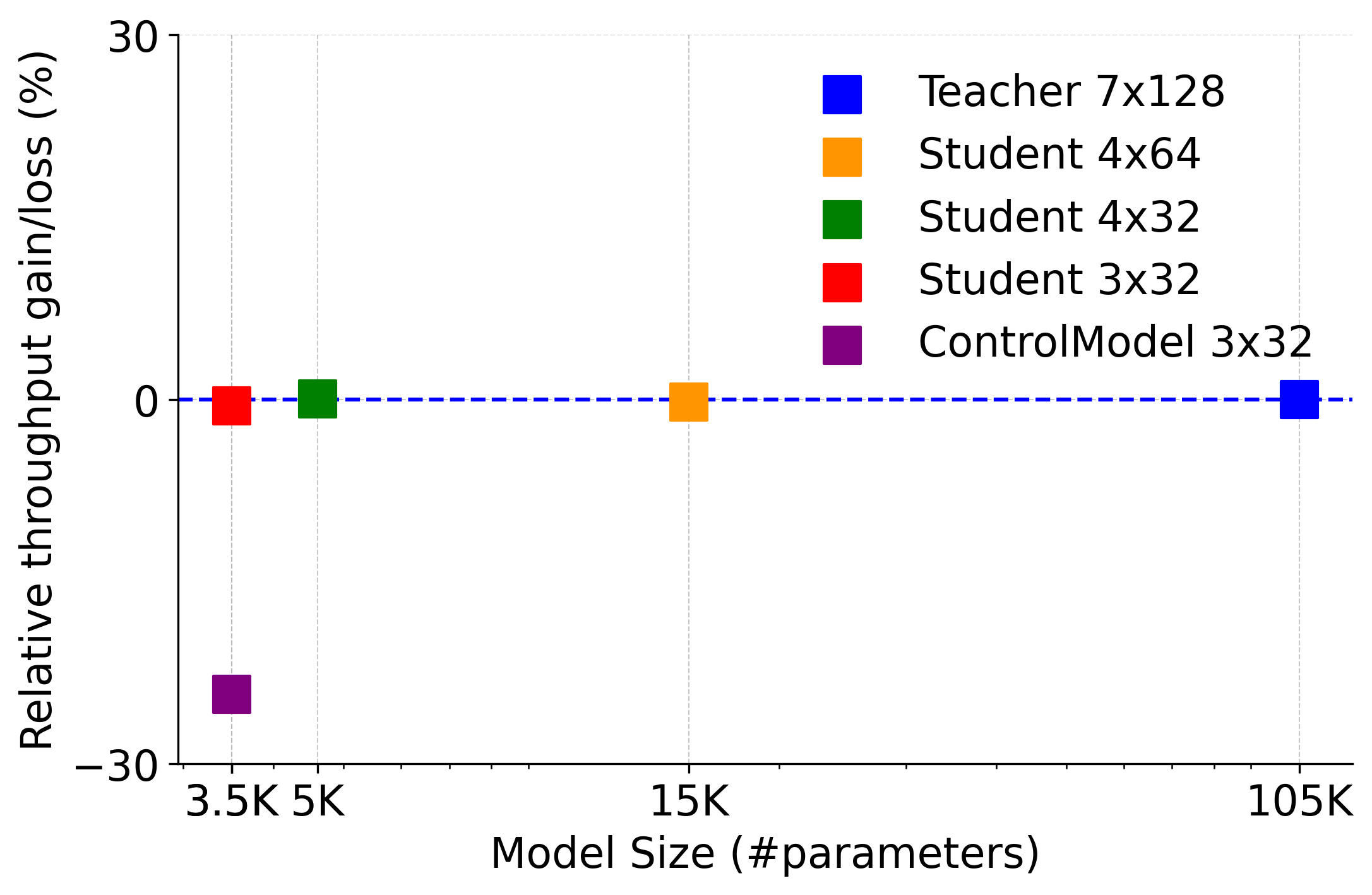}
        \caption{SCSU}
        \label{f}
    \end{subfigure}
    \caption{Comparison of the average throughput achieved by each student relative to the teacher in the \ac{MIMO}, \ac{mMIMO} and \ac{SCSU} benchmark scenarios.}
    \label{fig:cdf_avg_throughput}
\end{figure*}

\Cref{fig:cdf_avg_throughput} and \Cref{tab:student_performance} (single-policy rows) demonstrate that distilling a teacher model (7 layers of 128 units) into progressively smaller student models—from 4 layers of 64 units down to 3 layers of 32 units, representing approximately a 30-fold compression—can effectively preserve performance. Specifically, throughput degradation remains within $1\%$ across the MIMO, mMIMO, and SCSU scenarios; \ac{BLER} deviation is limited to within $\pm 7\%$ relative to the teacher model, with the mid-sized student model (4 layers of 32 units) even achieving a $6.6\%$ improvement in \ac{BLER} for SCSU. Moreover, the episodic reward declines by no more than $1.7\%$. In contrast, the control model (3 layers of 32 units), trained from scratch, significantly underperforms—by approximately $12\%$ in \ac{MIMO}, $10\%$ in \ac{mMIMO}, and $25\%$ in \ac{SCSU}—as indicated by the purple markers falling clearly below the 0\% reference line in \cref{fig:cdf_avg_throughput}.

\begin{figure*}[htbp]
    \centering
    \begin{subfigure}{0.3\textwidth}
        \centering
        \includegraphics[width=\linewidth]{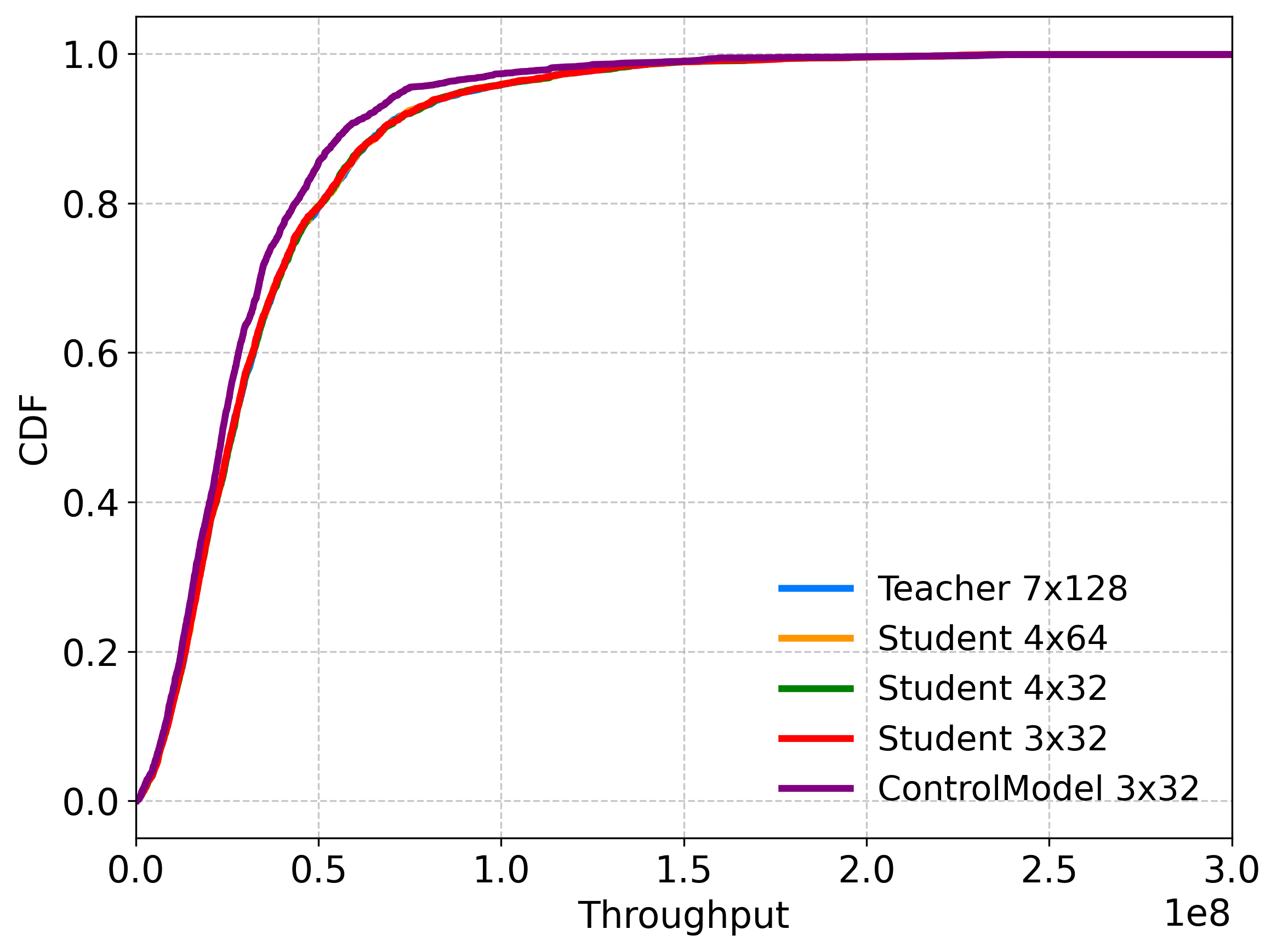}
        \caption{MIMO}
        \label{a}
    \end{subfigure}
    \hfill
    \begin{subfigure}{0.3\textwidth}
        \centering
        \includegraphics[width=\linewidth]{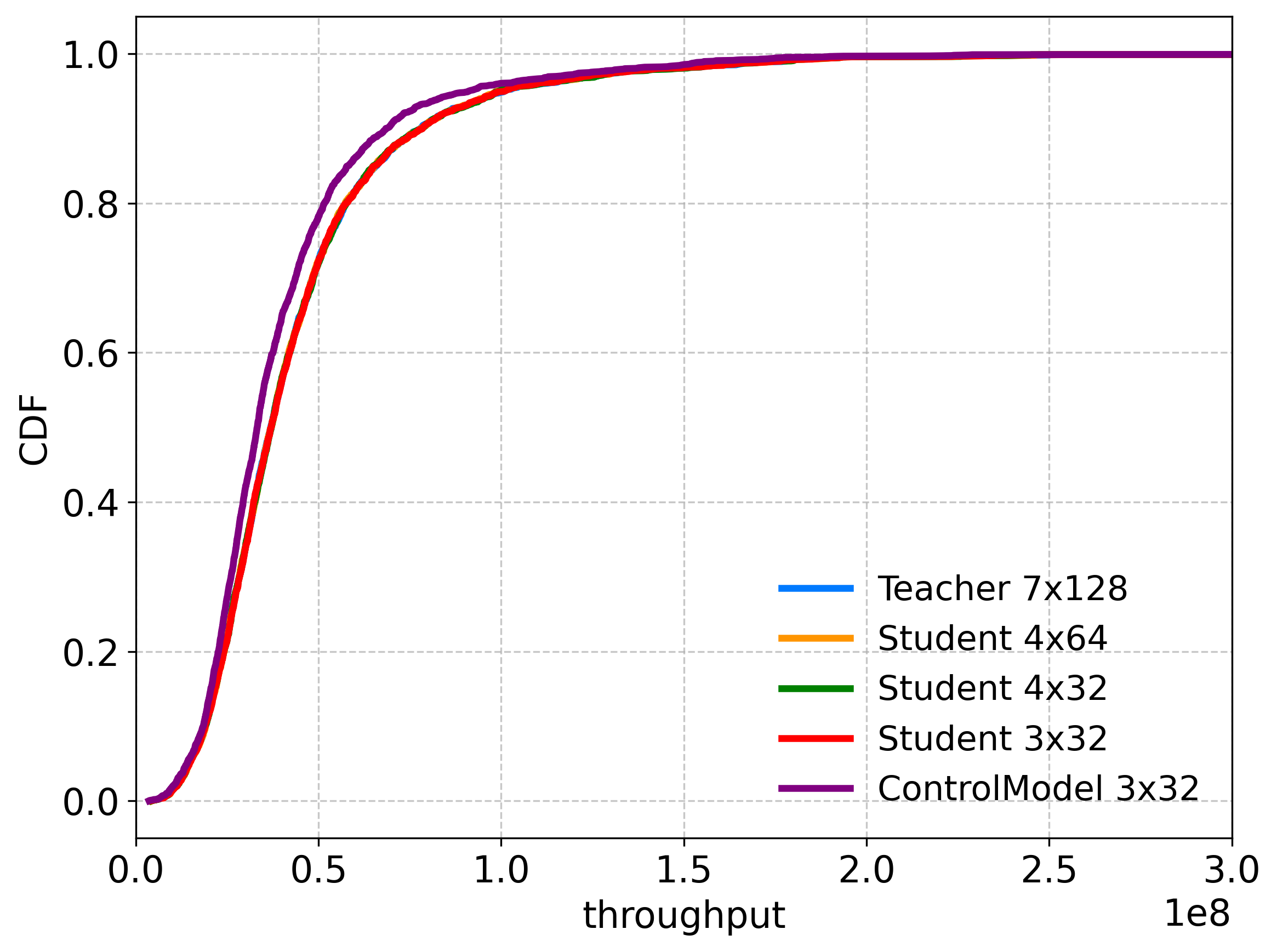}
        \caption{mMIMO}
        \label{b}
    \end{subfigure}
    \hfill
    \begin{subfigure}{0.3\textwidth}
        \centering
        \includegraphics[width=\linewidth]{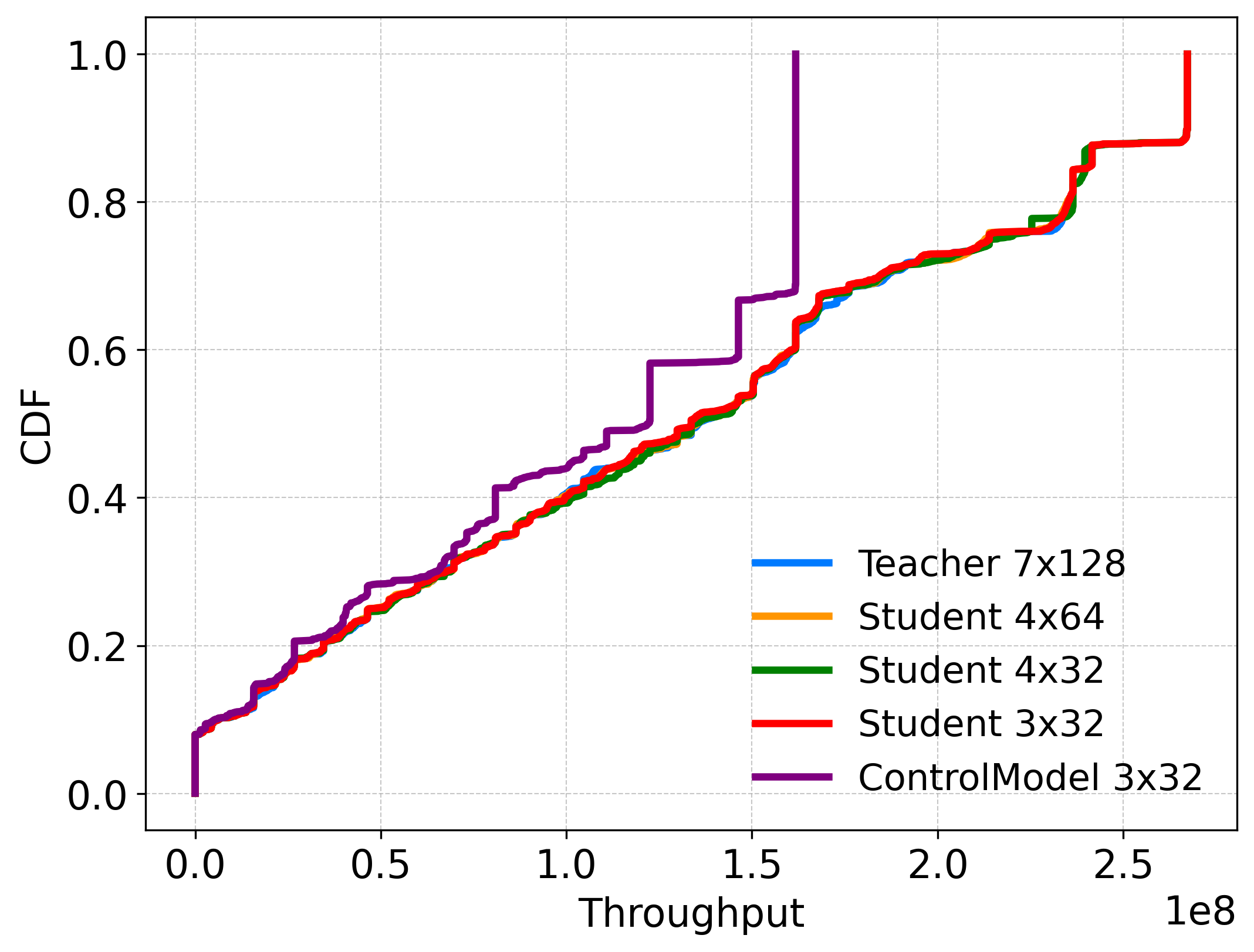}
        \caption{SCSU}
        \label{c}
    \end{subfigure}
    
    \vspace{0.5cm}
    
    \begin{subfigure}{0.3\textwidth}
        \centering
        \includegraphics[width=\linewidth]{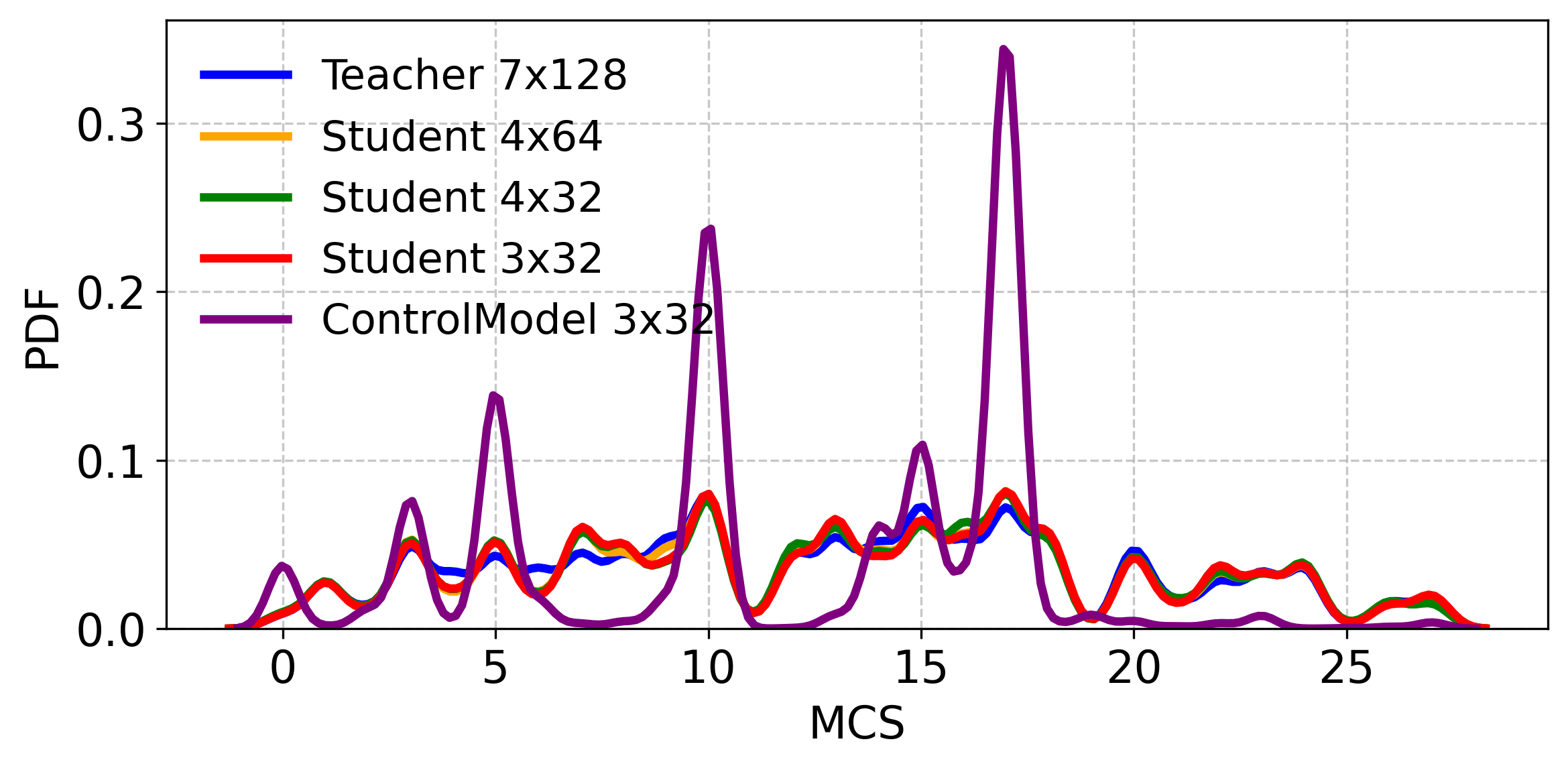}
        \caption{MIMO}
    \end{subfigure}
    \hfill
    \begin{subfigure}{0.3\textwidth}
        \centering
        \includegraphics[width=\linewidth]{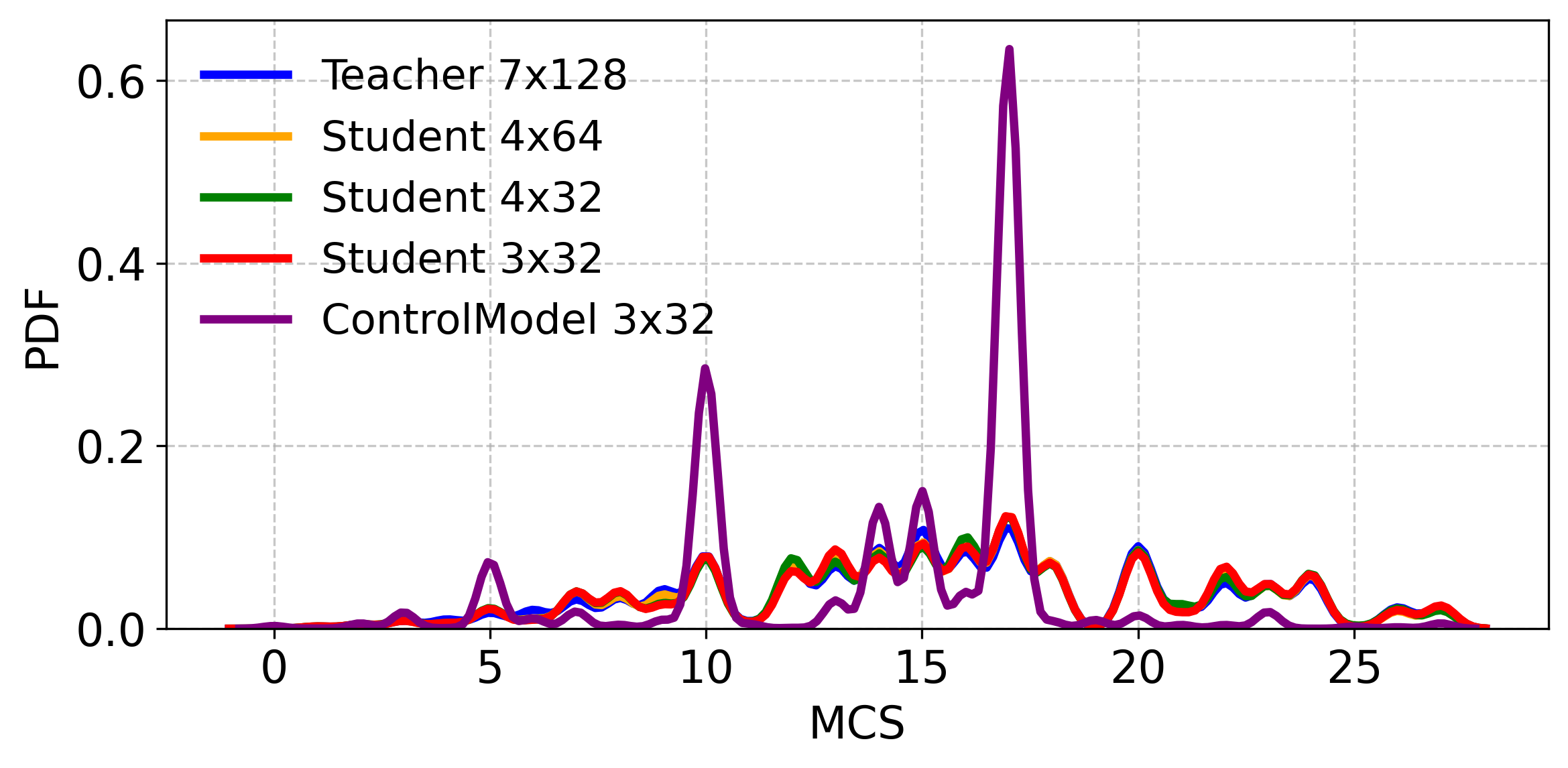}
        \caption{mMIMO}
    \end{subfigure}
    \hfill
    \begin{subfigure}{0.3\textwidth}
        \centering
        \includegraphics[width=\linewidth]{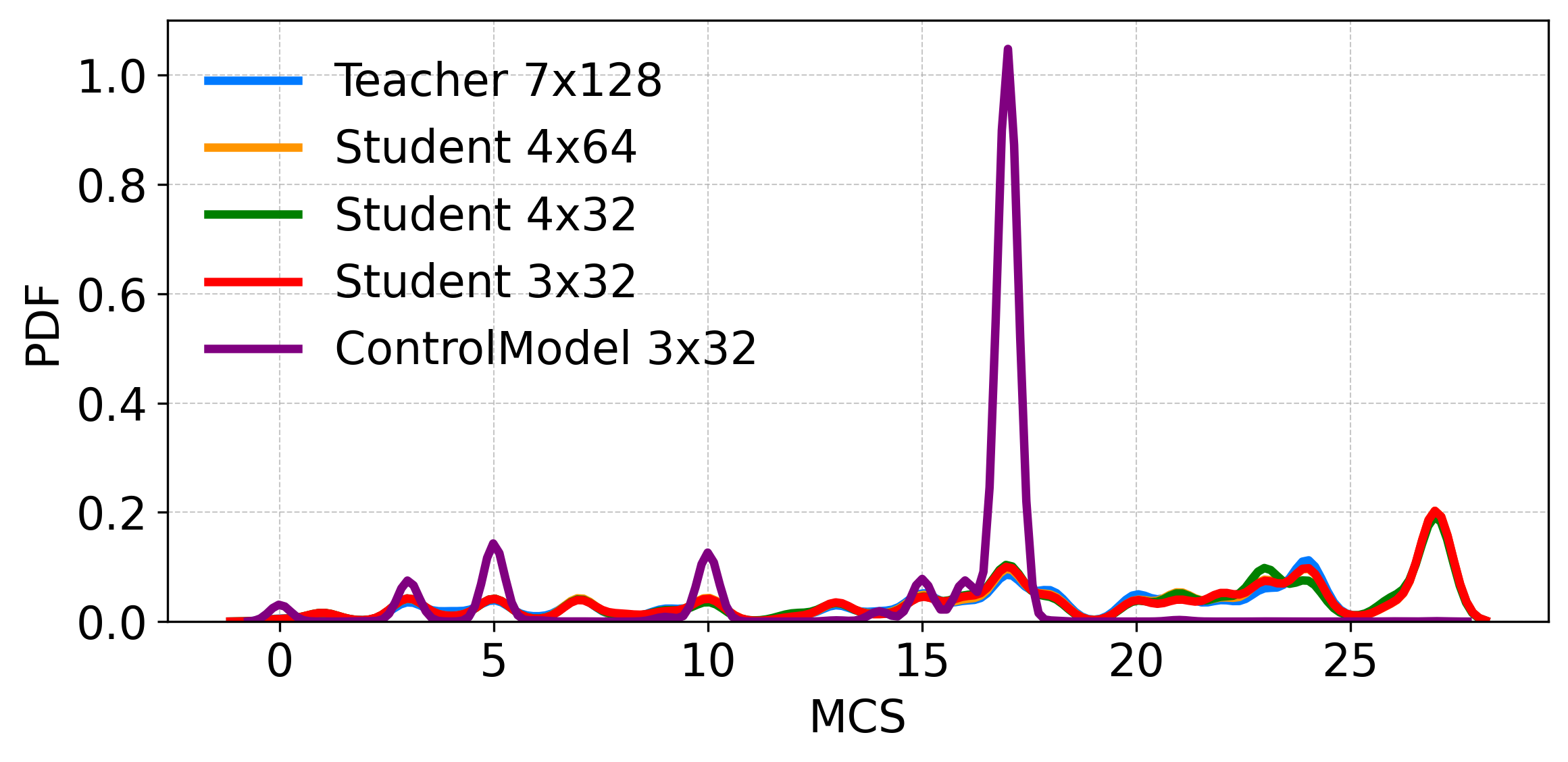}
        \caption{SCSU}
    \end{subfigure}
    \caption{Comparison of the teacher and students policies across the three benchmark scenarios in terms of:  \ac{CDF} of the \ac{UE} throughput (a)-(c); and actions distribution, represented by the \ac{PDF} of selected \ac{MCS} values (d)-(f).}
    \label{fig:pdf_mcs}
\end{figure*}

\Cref{fig:pdf_mcs} further confirms that the distilled student models effectively replicate the teacher model’s behavior, both in terms of throughput and the actions selected. Specifically, \cref{fig:pdf_mcs}(a)--(c) illustrate a near-perfect overlap between the teacher (blue) and student models (orange, green, red) in the \ac{CDF} of \ac{UE} throughput. In contrast, the smaller control model trained from scratch (purple) deviates considerably—especially in the \ac{SCSU} scenario. Similarly, \cref{fig:pdf_mcs}(d)--(f) confirm that the student models closely match the teacher’s distribution of \ac{MCS} indices. The control model, however, collapses onto a limited subset of indices, displaying pronounced peaks near indices 10 and 16, and rarely selecting indices above 19.

Collectively, these results confirm that single-policy distillation, when combined with domain randomization, effectively preserves the teacher model’s generalization capability, even after substantial compression—approximately 30-fold. Directly training a similarly sized model, however, leads to poor generalization performance.

\subsection{Generalization via Multi-Policy Distillation}

\Cref{tab:student_performance} (multi-policy rows) shows that a single student model, distilled from the combined knowledge of three scenario-specific teacher models, closely retains the performance of the original experts despite an approximate 30-fold parameter reduction (to a model with 3 layers of 32 units). Across the three benchmark environments, the throughput deficit of the smallest student model never exceeds $-2.8\%$ (MIMO scenario), while the larger student models maintain throughput within $-1.0\%$ to $-2.3\%$. Changes in \ac{BLER} remain modest, ranging from an improvement of $-4.0\%$ (mMIMO scenario) to a maximum increase of $+5.7\%$ (MIMO scenario), never exceeding a +6\% deviation. The maximum observed reduction in episodic reward is $3.7\%$. Collectively, these tight performance bounds indicate that multi-policy distillation effectively compresses and integrates specialized knowledge from multiple scenario-specific expert models into a single lightweight generalist model capable of robustly generalizing across diverse benchmark conditions.

\section{Conclusions}\label{sec:7}

In this paper, we demonstrated that policy distillation is a viable approach to replacing traditional \ac{RAN} functions with \ac{AI}-based counterparts that conform to the computational and memory constraints of existing baseband hardware. Focusing on the \ac{LA} function, we applied domain randomization to assemble richly varied training sets from multiple simulated \ac{RAN} scenarios. These datasets enabled us to train lightweight student models that faithfully imitate the decision logic of high-capacity teacher models. We assessed two distillation paradigms: (i) single-policy distillation, which compresses a unified, scenario-agnostic teacher into a compact student; and (ii) multi-policy distillation, which merges the expertise of several scenario-specific teachers into a single generalist student. Through experiments, we showed that both strategies substantially reduce model size—bringing their footprints well within existing hardware limits—while retaining near-teacher performance and broad generalization across diverse network conditions. Our findings underscore policy distillation as a practical, minimally invasive pathway for embedding AI-driven \ac{RAN} functions into current deployments.

\bibliographystyle{IEEEtran}
\balance
\bibliography{references.bib}

\end{document}